\documentclass{llncs} 
\usepackage{graphicx}
\usepackage{color}
\usepackage{array}
\usepackage{amsmath} 
\usepackage{amssymb}
\usepackage{booktabs}
\usepackage{subcaption}
\newcommand{\refsubfig}[2]{\ref{#1}(\subref{#2})}

\begin{document} 

\sloppy 
\newtheorem{defi}{Definition}
\newtheorem{prop}[defi]{Proposition}
\newtheorem{lemm}[defi]{Lemma}
\newtheorem{theo}[defi]{Theorem}
\newtheorem{coro}[defi]{Corollary}
\newtheorem{exam}[defi]{Example}
\newtheorem{fact}[defi]{Fact}
\newtheorem{algo}[defi]{Algorithm}

\newcommand{\sgn}{\mathrm{sgn}}
\newcommand{\define}{\stackrel{\mathrm{def}}{=}}
\newcommand{\argmin}{\mathop{\mathrm{argmin\,}}}
\newcommand{\argmax}{\mathop{\mathrm{argmax\,}}}
\newcommand{\rank}[1]{\mathrm{rank}\left(#1\right)}
\newcommand{\tr}[1]{\mathrm{tr}(#1)}
\newcommand{\diag}[1]{\mathrm{diag}\left(#1\right)}
\newcommand{\sinc}{\mathrm{sinc}\;}
\newcommand{\unorm}[1]{\|#1\|}
\newcommand{\unorms}[1]{\unorm{#1}^2}
\newcommand{\inner}[2]{\langle #1,#2\rangle}
\newcommand{\iid}{\stackrel{\mathrm{i.i.d.}}{\sim}}
\newcommand{\indep}{\mathop{\perp\!\!\!\perp}}

%%%%%%%%%%%%%%%%%%%%%%%%%%%%%%%%%%%%%%%%%%%%%%%%%%%%55

\newcommand{\mathbbC}{\mathbb{C}}
\newcommand{\mathbbE}{\mathbb{E}}
\newcommand{\mathbbN}{\mathbb{N}}
\newcommand{\mathbbR}{\mathbb{R}}
\newcommand{\mathbbS}{\mathbb{S}}

\newcommand{\boldzero}{{\boldsymbol{0}}}
\newcommand{\boldone}{{\boldsymbol{1}}}

\newcommand{\boldA}{{\boldsymbol{A}}}
\newcommand{\boldB}{{\boldsymbol{B}}}
\newcommand{\boldC}{{\boldsymbol{C}}}
\newcommand{\boldD}{{\boldsymbol{D}}}
\newcommand{\boldE}{{\boldsymbol{E}}}
\newcommand{\boldF}{{\boldsymbol{F}}}
\newcommand{\boldG}{{\boldsymbol{G}}}
\newcommand{\boldH}{{\boldsymbol{H}}}
\newcommand{\boldI}{{\boldsymbol{I}}}
\newcommand{\boldJ}{{\boldsymbol{J}}}
\newcommand{\boldK}{{\boldsymbol{K}}}
\newcommand{\boldL}{{\boldsymbol{L}}}
\newcommand{\boldM}{{\boldsymbol{M}}}
\newcommand{\boldN}{{\boldsymbol{N}}}
\newcommand{\boldO}{{\boldsymbol{O}}}
\newcommand{\boldP}{{\boldsymbol{P}}}
\newcommand{\boldQ}{{\boldsymbol{Q}}}
\newcommand{\boldR}{{\boldsymbol{R}}}
\newcommand{\boldS}{{\boldsymbol{S}}}
\newcommand{\boldT}{{\boldsymbol{T}}}
\newcommand{\boldU}{{\boldsymbol{U}}}
\newcommand{\boldV}{{\boldsymbol{V}}}
\newcommand{\boldW}{{\boldsymbol{W}}}
\newcommand{\boldX}{{\boldsymbol{X}}}
\newcommand{\boldY}{{\boldsymbol{Y}}}
\newcommand{\boldZ}{{\boldsymbol{Z}}}

\newcommand{\bolda}{{\boldsymbol{a}}}
\newcommand{\boldb}{{\boldsymbol{b}}}
\newcommand{\boldc}{{\boldsymbol{c}}}
\newcommand{\boldd}{{\boldsymbol{d}}}
\newcommand{\bolde}{{\boldsymbol{e}}}
\newcommand{\boldf}{{\boldsymbol{f}}}
\newcommand{\boldg}{{\boldsymbol{g}}}
\newcommand{\boldh}{{\boldsymbol{h}}}
\newcommand{\boldi}{{\boldsymbol{i}}}
\newcommand{\boldj}{{\boldsymbol{j}}}
\newcommand{\boldk}{{\boldsymbol{k}}}
\newcommand{\boldl}{{\boldsymbol{l}}}
\newcommand{\boldm}{{\boldsymbol{m}}}
\newcommand{\boldn}{{\boldsymbol{n}}}
\newcommand{\boldo}{{\boldsymbol{o}}}
\newcommand{\boldp}{{\boldsymbol{p}}}
\newcommand{\boldq}{{\boldsymbol{q}}}
\newcommand{\boldr}{{\boldsymbol{r}}}
\newcommand{\bolds}{{\boldsymbol{s}}}
\newcommand{\boldt}{{\boldsymbol{t}}}
\newcommand{\boldu}{{\boldsymbol{u}}}
\newcommand{\boldv}{{\boldsymbol{v}}}
\newcommand{\boldw}{{\boldsymbol{w}}}
\newcommand{\boldx}{{\boldsymbol{x}}}
\newcommand{\boldy}{{\boldsymbol{y}}}
\newcommand{\boldz}{{\boldsymbol{z}}}

\newcommand{\boldalpha}{{\boldsymbol{\alpha}}}
\newcommand{\boldbeta}{{\boldsymbol{\beta}}}
\newcommand{\boldgamma}{{\boldsymbol{\gamma}}}
\newcommand{\bolddelta}{{\boldsymbol{\delta}}}
\newcommand{\boldepsilon}{{\boldsymbol{\epsilon}}}
\newcommand{\boldzeta}{{\boldsymbol{\zeta}}}
\newcommand{\boldeta}{{\boldsymbol{\eta}}}
\newcommand{\boldtheta}{{\boldsymbol{\theta}}}
\newcommand{\boldiota}{{\boldsymbol{\iota}}}
\newcommand{\boldkappa}{{\boldsymbol{\kappa}}}
\newcommand{\boldlambda}{{\boldsymbol{\lambda}}}
\newcommand{\boldmu}{{\boldsymbol{\mu}}}
\newcommand{\boldnu}{{\boldsymbol{\nu}}}
\newcommand{\boldxi}{{\boldsymbol{\xi}}}
\newcommand{\boldpi}{{\boldsymbol{\pi}}}
\newcommand{\boldrho}{{\boldsymbol{\rho}}}
\newcommand{\boldsigma}{{\boldsymbol{\sigma}}}
\newcommand{\boldtau}{{\boldsymbol{\tau}}}
\newcommand{\boldupsilon}{{\boldsymbol{\upsilon}}}
\newcommand{\boldphi}{{\boldsymbol{\phi}}}
\newcommand{\boldchi}{{\boldsymbol{\chi}}}
\newcommand{\boldpsi}{{\boldsymbol{\psi}}}
\newcommand{\boldomega}{{\boldsymbol{\omega}}}
\newcommand{\boldGamma}{{\boldsymbol{\Gamma}}}
\newcommand{\boldDelta}{{\boldsymbol{\Delta}}}
\newcommand{\boldTheta}{{\boldsymbol{\Theta}}}
\newcommand{\boldLambda}{{\boldsymbol{\Lambda}}}
\newcommand{\boldXi}{{\boldsymbol{\Xi}}}
\newcommand{\boldPi}{{\boldsymbol{\Pi}}}
\newcommand{\boldSigma}{{\boldsymbol{\Sigma}}}
\newcommand{\boldUpsilon}{{\boldsymbol{\Upsilon}}}
\newcommand{\boldPhi}{{\boldsymbol{\Phi}}}
\newcommand{\boldPsi}{{\boldsymbol{\Psi}}}
\newcommand{\boldOmega}{{\boldsymbol{\Omega}}}
\newcommand{\boldvarepsilon}{{\boldsymbol{\varepsilon}}}
\newcommand{\boldvartheta}{{\boldsymbol{\vartheta}}}
\newcommand{\boldvarpi}{{\boldsymbol{\varpi}}}
\newcommand{\boldvarrho}{{\boldsymbol{\varrho}}}
\newcommand{\boldvarsigma}{{\boldsymbol{\varsigma}}}
\newcommand{\boldvarphi}{{\boldsymbol{\varphi}}}
 
\newcommand{\calA}{{\mathcal{A}}}
\newcommand{\calB}{{\mathcal{B}}}
\newcommand{\calC}{{\mathcal{C}}}
\newcommand{\calD}{{\mathcal{D}}}
\newcommand{\calE}{{\mathcal{E}}}
\newcommand{\calF}{{\mathcal{F}}}
\newcommand{\calG}{{\mathcal{G}}}
\newcommand{\calH}{{\mathcal{H}}}
\newcommand{\calI}{{\mathcal{I}}}
\newcommand{\calJ}{{\mathcal{J}}}
\newcommand{\calK}{{\mathcal{K}}}
\newcommand{\calL}{{\mathcal{L}}}
\newcommand{\calM}{{\mathcal{M}}}
\newcommand{\calN}{{\mathcal{N}}}
\newcommand{\calO}{{\mathcal{O}}}
\newcommand{\calP}{{\mathcal{P}}}
\newcommand{\calQ}{{\mathcal{Q}}}
\newcommand{\calR}{{\mathcal{R}}}
\newcommand{\calS}{{\mathcal{S}}}
\newcommand{\calT}{{\mathcal{T}}}
\newcommand{\calU}{{\mathcal{U}}}
\newcommand{\calV}{{\mathcal{V}}}
\newcommand{\calW}{{\mathcal{W}}}
\newcommand{\calX}{{\mathcal{X}}}
\newcommand{\calY}{{\mathcal{Y}}}
\newcommand{\calZ}{{\mathcal{Z}}}

\newcommand{\thetah}{{\widehat{\theta}}}
\newcommand{\thetat}{{\widetilde{\theta}}}
\newcommand{\boldthetah}{{\widehat{\boldtheta}}}
\newcommand{\boldthetat}{{\widetilde{\boldtheta}}}
\newcommand{\boldzetah}{{\widehat{\boldzeta}}}
\newcommand{\calAh}{{\widehat{\calA}}}
\newcommand{\boldEh}{{\widehat{\boldE}}}
\newcommand{\boldGh}{{\widehat{\boldG}}}
\newcommand{\boldUh}{{\widehat{\boldU}}}
\newcommand{\Eh}{{\widehat{E}}}
\newcommand{\jh}{{\widehat{j}}}
\newcommand{\boldbetah}{{\widehat{\boldbeta}}}
\newcommand{\boldbetat}{{\widetilde{\boldbeta}}}

\newcommand{\mathrmd}{{\mathrm{d}}}

%%%%%%%%%%%%%%%%%%%%%%%%%%%%%%%%%%%%%%%%%%%%%%%%%%%%%%%

\newcommand{\domainx}{\calX}
\newcommand{\domainy}{\calY}
\newcommand{\domainz}{\calZ}
\newcommand{\inputdim}{d}
\newcommand{\reducedim}{m}
\newcommand{\reducedimh}{\widehat{\reducedim}}
\newcommand{\ntask}{m}
\newcommand{\nmix}{c}
\newcommand{\inputdimx}{\inputdim_{\mathrm{x}}}
\newcommand{\inputdimy}{\inputdim_{\mathrm{y}}}

\newcommand{\ratiosymbol}{r}
\newcommand{\ratio}{\ratiosymbol^\ast}
\newcommand{\ratioh}{\widehat{\ratiosymbol}}
\newcommand{\ratiomodel}{\ratiosymbol}

\newcommand{\densitysymbol}{p}
\newcommand{\cumulativesymbol}{P}
\newcommand{\density}{\densitysymbol}
\newcommand{\densitydet}{\widetilde{\densitysymbol}_{\mathrm{de}}^\ast}
\newcommand{\cumulativede}{\cumulativesymbol_{\mathrm{de}}^\ast}
\newcommand{\cumulativenu}{\cumulativesymbol_{\mathrm{nu}}^\ast}
\newcommand{\cumulativedeh}{\widehat{\cumulativesymbol}_{\mathrm{de}}}
\newcommand{\densityh}{\widehat{\densitysymbol}}
\newcommand{\densityt}{\widetilde{\densitysymbol}}

\newcommand{\densitymodel}{\densitysymbol}
\newcommand{\prior}{\pi}
\newcommand{\posterior}{\densitysymbol}
\newcommand{\evidence}{\densitysymbol}

\newcommand{\xy}{\mathrm{xy}}
\newcommand{\zy}{\mathrm{zy}}

\newcommand{\pxy}{p_{\mathrm{xy}}}
\newcommand{\pzy}{p_{\mathrm{zy}}}
\newcommand{\px}{p_{\mathrm{x}}}
\newcommand{\pz}{p_{\mathrm{z}}}
\newcommand{\py}{p_{\mathrm{y}}}
\newcommand{\boldxtr}{\boldx^{\mathrm{de}}}
\newcommand{\boldxte}{\boldx^{\mathrm{nu}}}
\newcommand{\xtr}{x^{\mathrm{de}}}
\newcommand{\xte}{x^{\mathrm{nu}}}
\newcommand{\ytr}{y^{\mathrm{de}}}
\newcommand{\yte}{y^{\mathrm{nu}}}
\newcommand{\epsilontr}{\epsilon^{\mathrm{de}}}
\newcommand{\epsilonte}{\epsilon^{\mathrm{nu}}}
\newcommand{\ntr}{n_{\mathrm{de}}}
\newcommand{\nte}{n_{\mathrm{nu}}}
\newcommand{\densitytr}{\densitysymbol_{\mathrm{de}}^\ast}
\newcommand{\densityte}{\densitysymbol_{\mathrm{nu}}^\ast}
\newcommand{\Xtr}{X^{\mathrm{de}}}
\newcommand{\Xte}{X^{\mathrm{nu}}}
\newcommand{\boldXtr}{\boldX^{\mathrm{de}}}
\newcommand{\boldXte}{\boldX^{\mathrm{nu}}}
\newcommand{\boldytr}{\boldy^{\mathrm{de}}}
\newcommand{\boldyte}{\boldy^{\mathrm{nu}}}
\newcommand{\Gen}{G}
\newcommand{\Genh}{\widehat{\Gen}}

\newcommand{\boldxnu}{\boldx^{\mathrm{nu}}}
\newcommand{\boldxde}{\boldx^{\mathrm{de}}}
\newcommand{\boldxt}{\widetilde{\boldx}}

\newcommand{\boldunu}{\boldu^{\mathrm{nu}}}
\newcommand{\boldude}{\boldu^{\mathrm{de}}}
\newcommand{\boldvnu}{\boldv^{\mathrm{nu}}}
\newcommand{\boldvde}{\boldv^{\mathrm{de}}}
\newcommand{\boldunuh}{\widehat{\boldu}^{\mathrm{nu}}}
\newcommand{\boldudeh}{\widehat{\boldu}^{\mathrm{de}}}
\newcommand{\unu}{u^{\mathrm{nu}}}
\newcommand{\ude}{u^{\mathrm{de}}}

\newcommand{\nde}{n_{\mathrm{de}}}
\newcommand{\nnu}{n_{\mathrm{nu}}}
\newcommand{\nsample}{n}

\newcommand{\nparam}{b}
\newcommand{\nparamm}{t}
\newcommand{\nclass}{c}
\newcommand{\ndet}{\widetilde{n}_{\mathrm{de}}}

\newcommand{\data}{D}
\newcommand{\datanu}{D^{\mathrm{nu}}}
\newcommand{\datade}{D^{\mathrm{de}}}

\newcommand{\diffsymbol}{f}
\newcommand{\diff}{\diffsymbol}
\newcommand{\diffh}{\widehat{\diffsymbol}} 
\newcommand{\diffmodel}{g}

\newcommand{\boldratiosymbol}{\boldr}
\newcommand{\boldratio}{\boldratiosymbol^\ast}
\newcommand{\boldratiomodel}{{\boldratiosymbol}}
\newcommand{\boldratiode}{\boldratiosymbol_{\mathrm{de}}^\ast}
\newcommand{\boldrationu}{\boldratiosymbol_{\mathrm{nu}}^\ast}
\newcommand{\boldratioh}{\widehat{\boldratiosymbol}}
\newcommand{\boldratiodeh}{\boldratioh_{\mathrm{de}}}
\newcommand{\boldrationuh}{\boldratioh_{\mathrm{nu}}}
\newcommand{\boldratiodemodel}{\boldratiomodel_{\mathrm{de}}}
\newcommand{\boldrationumodel}{\boldratiomodel_{\mathrm{nu}}}
\newcommand{\boldPhide}{\boldPhi_{\mathrm{de}}}
\newcommand{\boldPhinu}{\boldPhi_{\mathrm{nu}}}
\newcommand{\boldPside}{\boldPsi_{\mathrm{de}}}
\newcommand{\boldPsinu}{\boldPsi_{\mathrm{nu}}}

\newcommand{\Bregman}{\mathrm{BR}}
\newcommand{\fdivergence}{\mathrm{ASC}}

\newcommand{\function}{f^\ast}
\newcommand{\functionmodel}{f}
\newcommand{\functionh}{\widehat{f}}

\newcommand{\xde}{x^{\mathrm{de}}}
\newcommand{\xdet}{\widetilde{x}^{\mathrm{de}}}
\newcommand{\xnu}{x^{\mathrm{nu}}}
\newcommand{\zde}{z^{\mathrm{de}}}
\newcommand{\boldxdet}{\widetilde{\boldx}^{\mathrm{de}}}
\newcommand{\boldxnut}{\widetilde{\boldx}^{\mathrm{nu}}}
\newcommand{\etade}{\eta^{\mathrm{de}}}
\newcommand{\etanu}{\eta^{\mathrm{nu}}}
\newcommand{\boldepsilonde}{\boldepsilon^{\mathrm{de}}}
\newcommand{\boldepsilondet}{\widetilde{\boldepsilon}^{\mathrm{de}}}
\newcommand{\epsilonde}{\epsilon^{\mathrm{de}}}
\newcommand{\epsilonnu}{\epsilon^{\mathrm{nu}}}
\newcommand{\epsilondeh}{\widehat{\epsilon}^{\mathrm{de}}}
\newcommand{\epsilondet}{\widetilde{\epsilon}^{\mathrm{de}}}
\newcommand{\boldrde}{\boldr^{\mathrm{de}}}
\newcommand{\boldrnu}{\boldr^{\mathrm{nu}}}
\newcommand{\Ade}{A^{\mathrm{de}}}
\newcommand{\Anu}{A^{\mathrm{nu}}}
\newcommand{\Xde}{X^{\mathrm{de}}}
\newcommand{\Xdet}{\widetilde{X}^{\mathrm{de}}}
\newcommand{\Xnu}{X^{\mathrm{nu}}}
\newcommand{\boldXde}{\boldX^{\mathrm{de}}}
\newcommand{\boldXdet}{\widetilde{\boldX}^{\mathrm{de}}}
\newcommand{\boldXnu}{\boldX^{\mathrm{nu}}}
\newcommand{\boldGde}{\boldG^{\mathrm{de}}}
\newcommand{\boldGnu}{\boldG^{\mathrm{nu}}}
\newcommand{\boldAde}{\boldA^{\mathrm{de}}}
\newcommand{\boldAnu}{\boldA^{\mathrm{nu}}}
\newcommand{\boldKdede}{\boldK_{\mathrm{de},\mathrm{de}}}
\newcommand{\boldKdenu}{\boldK_{\mathrm{de},\mathrm{nu}}}
\newcommand{\boldKde}{\boldK_{\mathrm{de}}}
\newcommand{\boldKnu}{\boldK_{\mathrm{nu}}}
\newcommand{\boldyde}{\boldy^{\mathrm{de}}}
\newcommand{\boldydet}{\widetilde{\boldy}^{\mathrm{de}}}
\newcommand{\Kdede}{K_{\mathrm{de},\mathrm{de}}}
\newcommand{\Kdenu}{K_{\mathrm{de},\mathrm{nu}}}
\newcommand{\Kde}{K_{\mathrm{de}}}
\newcommand{\yde}{y^{\mathrm{de}}}
\newcommand{\ydet}{\widetilde{y}^{\mathrm{de}}}
\newcommand{\ynu}{y^{\mathrm{nu}}}
\newcommand{\yh}{\widehat{y}}
\newcommand{\ynuh}{\widehat{y}^{\mathrm{nu}}}
\newcommand{\calXde}{\calX^{\mathrm{de}}}
\newcommand{\calYde}{\calY^{\mathrm{de}}}
\newcommand{\calZde}{\calZ^{\mathrm{de}}}
\newcommand{\calXnu}{\calX^{\mathrm{nu}}}
\newcommand{\calYnu}{\calY^{\mathrm{nu}}}
\newcommand{\calZnu}{\calZ^{\mathrm{nu}}}
\newcommand{\calZt}{\widetilde{\calZ}}

\newcommand{\Pb}{{P^{\ast}}'}
\newcommand{\Pa}{P^\ast}
\newcommand{\pb}{{p^{\ast}}'}
\newcommand{\pa}{p^\ast}
\newcommand{\nb}{n'}
\newcommand{\na}{n}
\newcommand{\calXb}{\calX'}
\newcommand{\calXa}{\calX}
\newcommand{\calXbt}{\widetilde{\calX}'}
\newcommand{\calXat}{\widetilde{\calX}}
\newcommand{\boldxb}{\boldx'}
\newcommand{\boldxa}{\boldx}
\newcommand{\boldHhde}{{\widehat{\boldH}}}
\newcommand{\boldhhde}{{\widehat{\boldh}}'}
\newcommand{\boldhhnu}{{\widehat{\boldh}}}
\newcommand{\boldGhde}{{\widehat{\boldG}}}
\newcommand{\boldghde}{{\widehat{\boldg}}'}
\newcommand{\boldghnu}{{\widehat{\boldg}}}
\newcommand{\nn}{\overline{n}}

\newcommand{\boldTh}{{\widehat{\boldT}}}
\newcommand{\boldgh}{{\widehat{\boldg}}}
\newcommand{\boldHh}{{\widehat{\boldH}}}
\newcommand{\boldhh}{{\widehat{\boldh}}}
\newcommand{\boldBh}{{\widehat{\boldB}}}
\newcommand{\Hh}{{\widehat{H}}}
\newcommand{\hh}{{\widehat{h}}}
\newcommand{\boldHt}{{\widetilde{\boldH}}}
\newcommand{\boldht}{{\widetilde{\boldh}}}
\newcommand{\boldDt}{{\widetilde{\boldD}}}
\newcommand{\boldWt}{{\widetilde{\boldW}}}
\newcommand{\boldXt}{{\widetilde{\boldX}}}
\newcommand{\Wt}{{\widetilde{W}}}
\newcommand{\Dt}{{\widetilde{D}}}
\newcommand{\Qt}{{\widetilde{Q}}}
\newcommand{\Qh}{{\widehat{Q}}}
\newcommand{\Ht}{{\widetilde{H}}}
\newcommand{\Xt}{{\widetilde{X}}}
\newcommand{\Jh}{\widehat{J}}
\newcommand{\Jt}{\widetilde{J}}
\newcommand{\Vh}{\widehat{V}}
\newcommand{\Fh}{\widehat{F}}
\newcommand{\Gh}{\widehat{G}}
\newcommand{\KL}{\mathrm{KL}}
\newcommand{\KLh}{\widehat{\KL}}
\newcommand{\KLt}{\widetilde{\KL}}
\newcommand{\boldxih}{\widehat{\boldxi}}
\newcommand{\boldalphah}{\widehat{\boldalpha}}
\newcommand{\boldalphat}{\widetilde{\boldalpha}}
\newcommand{\alphah}{\widehat{\alpha}}
\newcommand{\sigmah}{{\widehat{\sigma}}}
\newcommand{\lambdah}{{\widehat{\lambda}}}

\newcommand{\boldSb}{\boldS^{\mathrm{b}}}
\newcommand{\boldSw}{\boldS^{\mathrm{w}}}
\newcommand{\boldWb}{\boldW^{\mathrm{b}}}
\newcommand{\boldWw}{\boldW^{\mathrm{w}}}
\newcommand{\Wb}{W^{\mathrm{b}}}
\newcommand{\Ww}{W^{\mathrm{w}}}
\newcommand{\boldSlb}{\boldS^{\mathrm{lb}}}
\newcommand{\boldSlw}{\boldS^{\mathrm{lw}}}
\newcommand{\boldWlb}{\boldW^{\mathrm{lb}}}
\newcommand{\boldWlw}{\boldW^{\mathrm{lw}}}
\newcommand{\Wlb}{W^{\mathrm{lb}}}
\newcommand{\Wlw}{W^{\mathrm{lw}}}
\newcommand{\weight}{w}
\newcommand{\weighth}{\widehat{w}}

\newcommand{\resample}{\zeta}
\newcommand{\CEVOW}{\textrm{CV}_\textrm{W}}
\newcommand{\PCEVOW}{\textrm{PCV}_\textrm{W}}
\newcommand{\CEVOO}{\textrm{CV}_\textrm{O}}
\newcommand{\PCEVOO}{\textrm{PCV}_\textrm{O}}
\newcommand{\FEVOW}{\textrm{FV}_\textrm{W}}
\newcommand{\PFEVOW}{\textrm{PFV}_\textrm{W}}
\newcommand{\FEBVOW}{\textrm{FBV}_\textrm{OW}}
\newcommand{\PFEBVOW}{\textrm{PFBV}_\textrm{OW}}
\newcommand{\ALICE}{\textrm{ALICE}}
\newcommand{\PALICE}{\textrm{PALICE}}

\newcommand{\model}{M}
\newcommand{\GhatAL}{\Genh^{(AL)}}
\newcommand{\GhatMS}{\Genh^{(MS)}}
\newcommand{\GhatuCEAL}{\Genh^{(EAL)}}
\newcommand{\GhatCEAL}{\Genh^{(nEAL)}}

\newcommand{\hatgn}{\widehat{g}_n}
\newcommand{\gnstar}{g_n^*}
\newcommand{\alzn}{\alpha_0^n}
\newcommand{\supp}{\mathrm{supp}}
\newcommand{\gzn}{\ratio}
\newcommand{\Gconv}{\mathcal{G}^{(\mathrm{c})}}

\newcommand{\dom}{\mathbf{dom}}

%%% Local Variables: 
%%% mode: latex
%%% TeX-master: "main"
%%% End: 

\mainmatter 

\title{Clustering Unclustered Data}
\subtitle{Unsupervised Binary Labeling of Two Datasets\\
  Having Different Class Balances}
\author{Marthinus Christoffel du Plessis \and Masashi Sugiyama}
\institute{
Department of Computer Science, Tokyo Institute of Technology, \\
2-12-1 O-okayama, Meguro-ku, Tokyo 152-8552, Japan,\\
%Department of Computer Science \\
%Tokyo Institute of Technology \\
\email{christo@sg.cs.titech.ac.jp, sugi@cs.titech.ac.jp}
}
\maketitle

\begin{abstract}
We consider the unsupervised learning problem of assigning labels to
unlabeled data.
A naive approach is to use clustering methods,
but this works well only when data is properly clustered and each
cluster corresponds to an underlying class.
In this paper, we first show that this unsupervised labeling problem
in balanced binary cases can be solved 
if two unlabeled datasets having different class balances are available.
More specifically, estimation of the \emph{sign}
of the difference between probability densities of two unlabeled
datasets gives the solution.
We then introduce a new method to directly estimate the sign of the
density difference without density estimation.
Finally, we demonstrate the usefulness of the proposed method against
several clustering methods
on various toy problems and real-world datasets.
\end{abstract}

\section{Introduction}
Gathering labeled data is expensive and time consuming
in many practical machine learning problems,
and therefore class labels are often absent.
In this paper, we consider the problem of \emph{labeling},
which is aimed at giving a label to each sample.
Labeling is similar to classification,
but it is slightly simpler than classification
because classes do not have to be specified.
That is, labeling just tries to split unlabeled samples into disjoint subsets, 
and class labels such as male/female or positive/negative are not assigned
to samples.

A naive approach to the labeling problem
is to use a clustering technique
which is aimed at assigning a label to each sample
of the dataset to divide the dataset into disjoint clusters.
The tacit assumption in clustering is that the clusters correspond to the underlying classes. 
However, this assumption is often violated in practical datasets,
for example, when clusters are not well separated
or a dataset exhibits within-class multimodality. 

\begin{figure}[t] 
        \centering
        \begin{subfigure}[b]{0.48\textwidth}
                \centering
                \includegraphics[width=\textwidth]{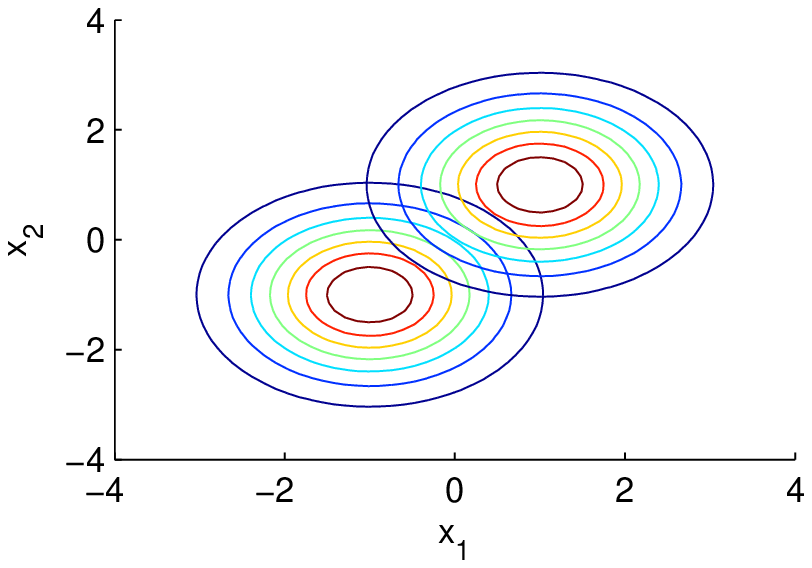} 
               \caption{Original distributions}
                \label{fig:distributions}
        \end{subfigure}\!\!\!  \\
        \begin{subfigure}[b]{0.48\textwidth}
                \centering
                \includegraphics[width=\textwidth]{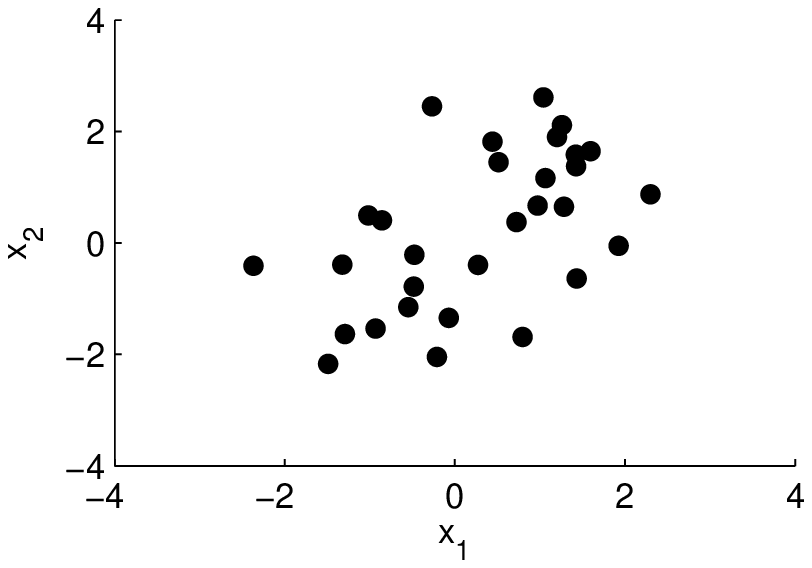} 
               \caption{Samples in $\calX_p$}
                \label{fig:Dataset1}
        \end{subfigure}\!\!\! 
        \begin{subfigure}[b]{0.48\textwidth}
                \centering
                \includegraphics[width=\textwidth]{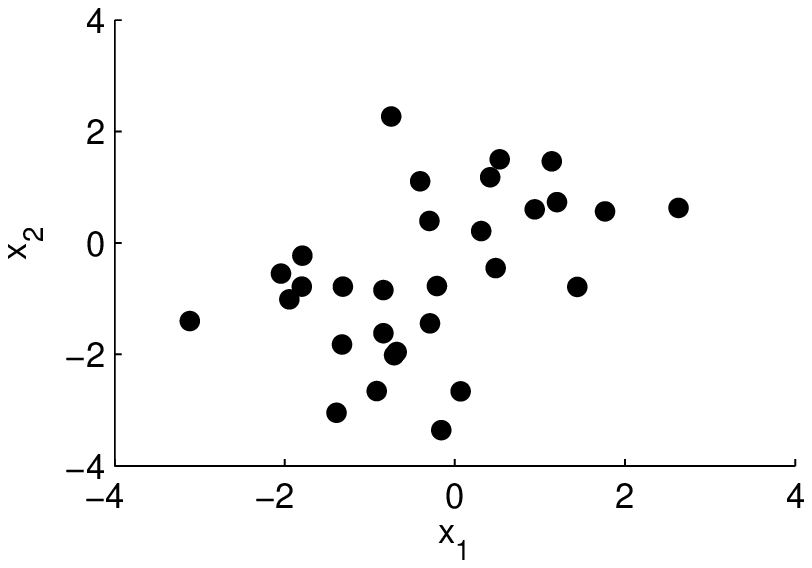} 
               \caption{Samples in $\calX_{p'}$}
                \label{fig:Dataset2}
        \end{subfigure}\!\!\!    \\
        \begin{subfigure}[b]{0.48\textwidth}
                \centering
                \includegraphics[width=\textwidth]{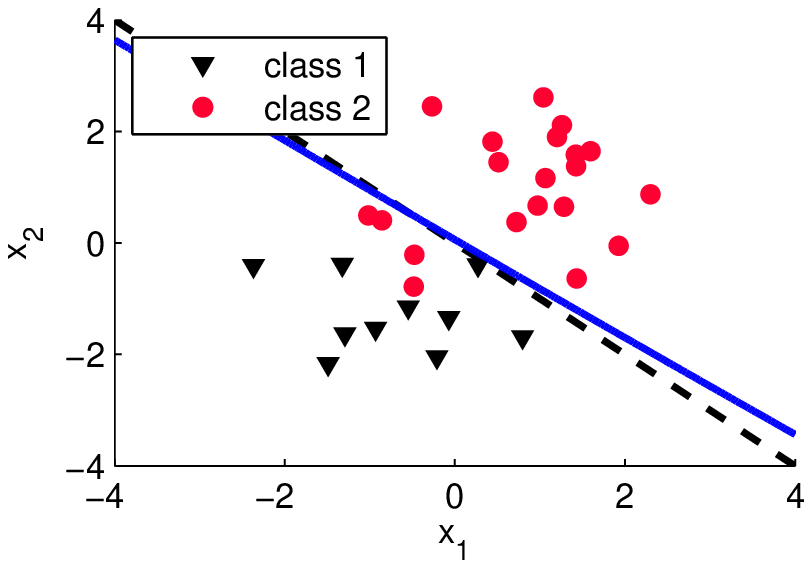} 
                \caption{Labeling of $\calX_p$}
                \label{fig:labellingXp} 
        \end{subfigure}\!\!\! 
        \begin{subfigure}[b]{0.48\textwidth}
                \centering
                \includegraphics[width=\textwidth]{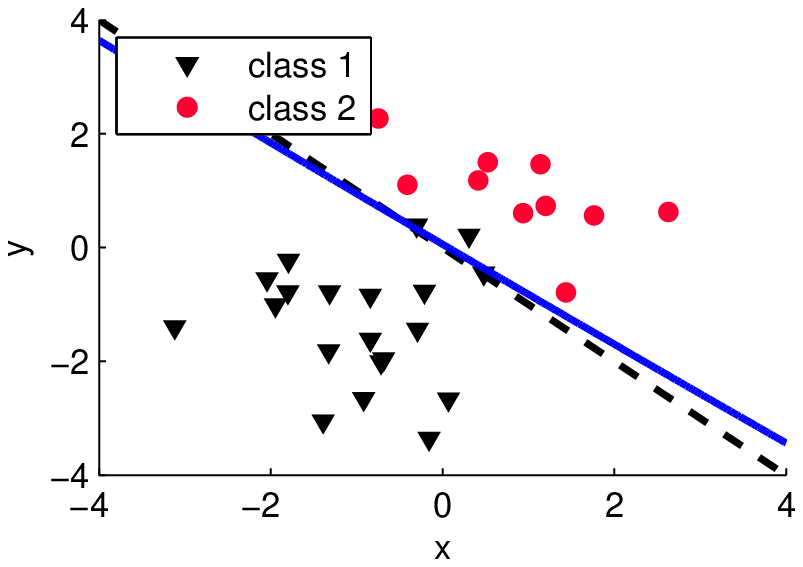}
                \caption{Labeling of $\calX_{p'}$}
                \label{fig:labellingXq} 
        \end{subfigure}\\
        \caption{Illustrative example of labeling samples from unbalanced datasets.
        Figures~\ref{fig:Dataset1} and \ref{fig:Dataset2} show the samples of the two datasets which differ only by class balance (the datasets are denoted as $\calX_p$ and $\calX_{p'}$).
        The discriminant estimated by the method that we propose in this paper 
		is given in blue and the optimal discriminant is given in the black dashed line. 
        The true underlying class labels (which are unknown) are illustrated in red and black.  }\label{fig:toyillustration}
\end{figure}

An example of the labeling problem is illustrated in Figure~\ref{fig:toyillustration}. Figure~\refsubfig{fig:toyillustration}{fig:distributions} denotes the densities of the two classes. 
Figure~\refsubfig{fig:toyillustration}{fig:Dataset1} denotes 
samples drawn from a mixture of the two original densities. 
Because the two clusters are highly overlapping, it may not be possible to properly label them by a clustering method. 

In this paper we show that if one more dataset with a different class balance is available (Figure~\refsubfig{fig:toyillustration}{fig:Dataset2}), the labeling problem can be solved 
(Figures~\refsubfig{fig:toyillustration}{fig:labellingXp} and \refsubfig{fig:toyillustration}{fig:labellingXq}).
More specifically, we show that
a labeling for the samples can be obtained by estimating \emph{the sign of the
difference between probability densities of two unlabeled datasets}.
A naive way is to first separately estimate
two densities from two sets of samples
and then take the sign of their difference to obtain a labeling.
However, this naive procedure violates \emph{Vapnik's principle}\cite{vapnik2000nature}: 
%
%
%
% We have two datasets (Figure~\refsubfig{fig:toyillustration}{fig:Dataset1} and Figure~\refsubfig{fig:toyillustration}{fig:Dataset2}) that differ by class balance. Because the two clusters are highly overlapping, it may not be possible to properly label them by a clustering method. Figures~\refsubfig{fig:toyillustration}{fig:labellingXp} and \refsubfig{fig:toyillustration}{fig:labellingXq} shows the optimal labeling and labeling by the method we 
% propose in this paper. By exploiting the class imbalance, we can find a labeling for the two datasets. 
 %
%
\begin{quote}
If you possess a restricted amount of
information for solving some problem, try
to solve the problem directly and never solve
a more general problem as an intermediate
step. It is possible that the available
information is sufficient for a direct solution
but is insufficient for solving a more general
intermediate problem.
\end{quote}
This principle was used in the development of \emph{support vector machines} (SVMs):
Rather than modeling two classes of samples,
SVM directly learns a decision boundary that is sufficient for performing
pattern recognition.

In the current context,
estimating two densities is more general than
labeling samples.
Thus, the above naive scheme may be improved by estimating
the density difference directly and then
taking its sign to obtain the class labels.
Recently, a method was introduced to 
directly estimate the density difference, 
called the \emph{least-squares density difference (LSDD)} estimator \cite{DensityDifferenceNIPS}.
Thus, the use of LSDD for labeling is expected to improve the performance.

However, the LSDD-based procedure is still indirect;
directly estimating the sign of the density difference
would be the most suitable approach to labeling.
In this paper, 
we show that the sign of the density difference can be directly estimated
by lower-bounding the $L_1$-distance between probability densities.
Based on this, we give a practical algorithm for labeling
and illustrate its usefulness 
through experiments on various real-world datasets.

% 
%
%In Section~\ref{sec:DirectSignDensityDifference}, we show that least-squares fitting can be 
%interpreted as the maximization of a lower-bound of the $L^2$ distance between probability 
%densities. By considering a more general set of distances, we show how the sign of the 
%density difference can be calculated directly. 
%In Section~\ref{sec:DirectSignDensityDifference}, we introduce a new method that {directly estimates the sign of the density difference}. 

\section{Problem Formulation and Fundamental Approaches}
\label{sec:ProblemFormulation}
In this section, we formulate the problem of labeling,
give our fundamental strategy, and consider
two naive approaches.

\subsection{Problem Formulation}
Suppose that there are two probability distributions
$\density(\boldx,y) $ and $\density'(\boldx, y)$
on $\boldx\in\mathbbR^d$ and $y\in\{1,-1\}$,
% \begin{align*}
% \left\{\left(\boldx_i, y_i\right) \right\}_{i=1}^{n} \iid \density(\boldx, y) 
% ~~\mbox{and}~~
% \left\{\left(\boldx_j', y_j'\right) \right\}_{j=1}^{n'} \iid \density'(\boldx, y).
% \end{align*}
which are different only in class balances:
\begin{align}
\density(y) \neq \density'(y) \qquad \mathrm{but } \qquad \density(\boldx | y) = \density'(\boldx|y).
\label{eq:ClassPriorChangeAssumption} 
\end{align}
From these distributions, 
we are given two sets of unlabeled samples:
\begin{align*}
\calX_\density = \{ \boldx_i \}_{i=1}^n \iid \density(\boldx) \:\: \textrm{and} \:\: \calX_{\density'} = \{ \boldx_j' \}_{j=1}^{n'} \iid \density'(\boldx).  
\end{align*}
The goal of labeling is
to obtain a labeling for the two sets of samples, $\calX_p$ and $\calX_{p'}$, that corresponds
to the underlying class labels $\{ y_i \}_{i=1}^n$ and $\{ y'_j \}_{j=1}^{n'}$.
However, different from classification,
we do not obtain correct class labels,
but we obtain correct class separation up to label commutation.

%
%\begin{align*}
%\calX = \left\{\boldx_i \right\}_{i=1}^n \:\:\textrm{and }\:\: \calX' = \left\{\boldx_j' \right\}_{j=1}^{n'}.
%\end{align*}
%\begin{align*} 
%\calX = \left\{\boldx_i \right\}_{i=1}^n \:\:\textrm{and }\:\: \calX' = \left\{\boldx_j' \right\}_{j=1}^{n'}
%\end{align*} 
%from the distributions $\density(\boldx)$ and $\density'(\boldx)$.

%We assume that we have unlabeled data from two distributions:
%\begin{align*}
%\calX_p = \left\{ \boldx_i \right\}_{i=1}^n \iid \density(\boldx) \: \textrm{and} \: \calX_{\density'} = \left\{ \boldx_j' \right\}_{j=1}^{n'} \iid \density'(\boldx).  
%\end{align*}
% We show in this section that we can obtain a labeling for samples from unbalanced datasets by estimating the sign of the density difference.

\subsection{Fundamental strategy}
We wish to obtain a labeling for samples in $\calX_p$ and $\calX_{p'}$. 
Here we show that we can obtain the solution for the case where the class priors are equal. 
%
%Since the class balance for both datasets are unknown, it is reasonable to perform classification assuming an
%equal class balance. 
We may write the class-posterior distribution for the equal prior case as
\begin{align*}
q(y=1|\boldx) & = \frac{p(\boldx|y)q(y)}{q(\boldx)},  
\end{align*}
where $q(y=1) = q(y=-1) = \frac{1}{2}$. A class label can then be assigned to a point by
evaluating 
\begin{align*}
\operatorname{sign}\left[  q(y=1|\boldx) - q(y=-1|\boldx)\right]
\end{align*} 
% 
%The class label can then be decided by
%\begin{align*}
%y = \begin{cases} 
%1 & q(y=1|\boldx) \geq q(y=-1|\boldx), \\
%-1 & \textrm{otherwise}.
%\end{cases}
%\end{align*}
We can write the criterion as 
\begin{align*}
q(y=1|\boldx) - q(y=-1|\boldx) & = \frac{p(\boldx|y=1)\frac{1}{2}}{q(\boldx)} - \frac{p(\boldx|y=-1)\frac{1}{2}}{q(\boldx)}, \\
& \propto  p(\boldx|y=1) - p(\boldx|y=-1).
\end{align*}
We do not have any labeled samples to calculate $p(\boldx|y=1) - p(\boldx|y=-1)$, but we can rewrite it in terms of marginal distributions. To see this, the above is multiplied with $p(y=1)-p'(y=1)$, which gives
\begin{align*}
p(\boldx|y=1) - p(\boldx|y=-1) & \propto \left[p(y=1) - p'(y=1) \right]\left[p(\boldx|y=1) - p(\boldx|y=-1) \right] \\
& \propto p(\boldx, y=1) - p'(\boldx, y=1)  \\
& \phantom{\propto} - p(y=1)p(\boldx|y=-1) + p'(y=1)p(\boldx|y=-1).
\end{align*}
Note that the sign may change since $p(y=1)-p'(y=1)$ may be positive or negative.  
To write the third and fourth term as a joint distribution, we add and subtract $p(\boldx|y=-1)$, giving
\begin{align*}
p(\boldx|y=1) - p(\boldx|y=-1) & \propto p(\boldx, y=1) - p'(\boldx, y=-1) +\left[1- p(y=1)\right]p(\boldx|y=\!-1) \\
& \phantom{\propto} \! - \left[1\! - p'(y=1)\right]p(\boldx|y=-1). 
\end{align*}
Since $p(y=-1) = 1-p(y=1)$ and $p'(y=-1) = 1-p'(y=1)$, we can express the above as
\begin{align*}
q(y=1|\boldx) - q(y=-1|\boldx) & \propto p(\boldx) - p'(\boldx).
\end{align*}
The exact class labels can not be recovered since the term $\density(y=1) - \density'(y=1)$ 
can be positive or negative. Therefore, we assign the label $y \in \left\{ 1,-1\right\}$ to a point $\boldx$ according to the following criterion:
\begin{align}
y = \operatorname{sign}{\left[\density(\boldx) - \density'(\boldx)  \right]}.
\label{eq:LabelingCriterion} 
\end{align}
Thus, now we need a good method to estimate
$\operatorname{sign}{\left[\density(\boldx) - \density'(\boldx)  \right]}$.

\subsection{Kernel Density Estimation}
\label{sec:DensityDiffViaKDE}
A naive approach to estimating the sign of density-difference is
to use \emph{kernel density estimators} (KDEs)
\cite{book:Silverman:1986}.
For Gaussian kernels, the KDE solutions are given by
\begin{align*}
% \widehat{\density}(\boldx)&:=\frac{1}{\nsample(2\pi\sigma^2)^{\inputdim/2}}\sum_{i=1}^{\nsample}
%   \exp\left(-\frac{\|\boldx-\boldx_i\|^2}{2\sigma^2}\right),\\
%   % second line
% \widehat{\density}'(\boldx)&:=\frac{1}{n'(2\pi\sigma'^2)^{\inputdim/2}}\sum_{j=1}^{\nsample'}
%   \exp\left(-\frac{\|\boldx-\boldx'_{j}\|^2}{2\sigma'^2}\right).
\widehat{\density}(\boldx)\propto\sum_{i=1}^{\nsample}
  \exp\left(-\frac{\|\boldx-\boldx_i\|^2}{2\sigma^2}\right)
~~\mbox{and}~~
  % second line
\widehat{\density}'(\boldx)\propto\sum_{j=1}^{\nsample'}
  \exp\left(-\frac{\|\boldx-\boldx'_{j}\|^2}{2\sigma'^2}\right).
\end{align*}
The Gaussian widths $\sigma$ and $\sigma'$ may be determined
based on least-squares cross-validation \cite{book:Haerdle+etal:2004}.
Finally, a labeling is obtained as
\begin{align}
y = \operatorname{sign}{\left[\widehat{\density}(\boldx) - \widehat{\density}'(\boldx)  \right]}.
\label{eq:LabelingCriterion} 
\end{align}

% However, KDE-based density-difference estimator may not be
% the best approach because of its two-step nature:
% Small estimation error incurred in each density estimate
% can cause a big error in the final density-difference estimate.
% More intuitively, good density estimators tend to be smooth
% and thus a density-difference estimator
% obtained from such smooth density estimators tends to be over-smoothed
% \cite{Biometrika:Hall+Wand:1988,JMA:Anderson+etal:1994}.

\subsection{Direct Estimation of the Density Difference}
\label{sec:DirectDensityDiffEstimation}
KDE is a nice density estimator,
but it is not necessarily suitable in density-difference estimation,
because small estimation error incurred in each density estimate
can cause a big error in the final density-difference estimate.
More intuitively, good density estimators tend to be smooth
and thus a density-difference estimator
obtained from such smooth density estimators tends to be over-smoothed
\cite{Biometrika:Hall+Wand:1988,JMA:Anderson+etal:1994}.

The density difference can be estimated in a single shot using the \emph{least-squares density difference} (LSDD) approach \cite{DensityDifferenceNIPS}.
In this approach, we directly fit a model $g(\boldx)$ to the density difference under the
square loss: 
%This can be done by fitting a model 
%$g(\boldx)$ to the density difference $\density(\boldx) - \density'(\boldx)$ under the square loss:
\begin{align*}
\widehat{g}=\argmin_{g} \frac{1}{2}\int{\left(g(\boldx) - (\density(\boldx) - \density'(\boldx)) \right)^2\mathrm{d}\boldx },
\end{align*}
which can be efficiently obtained for a kernel density-difference model.
A comprehensive review of LSDD is provided in Appendix~\ref{sec:AppendixLSDD}.
Finally, a labeling is obtained as
\begin{align*}
y = \operatorname{sign}[\widehat{g}(\boldx)].
%\label{eq:LabelingCriterion} 
\end{align*}

\section{Direct Estimation of the Sign of the Density Difference}
\label{sec:DirectEstimationSign}
We expect that an improved solution can be obtained by LSDD over KDEs
due to more direct nature of LSDD.
However, LSDD is still indirect
because the sign of density difference is inspected 
after the density difference is estimated.
In this section, we show how to directly estimate 
the sign of the density difference.

\subsection{Derivation of the Objective Function}
By lower-bounding the $L_1$-distance between probability densities, defined as 
\begin{align}
\int{\left|p(\boldx) - p'(\boldx) \right|\mathrm{d}\boldx}, 
\label{eq:L1Distance}
\end{align} 
we can obtain the sign of the density difference.
% an estimate for the class labels as an intermediate step. 
We begin by considering the following
self-evident relation:
\begin{align*}
|t| \geq tz , \:\:\textrm{if}\:\: |z| \leq 1.
\end{align*}
We can apply this relation at each point $\boldx$, to obtain
\begin{align*}
\left|\density(\boldx) - \density'(\boldx) \right| \geq g(\boldx)\left[\density(\boldx) - \density'(\boldx)  \right] \:\: \textrm{if}\:\: |g(\boldx)| \leq 1, \:\: \forall \boldx.
\end{align*}
By applying the above inequality to Eq.\eqref{eq:L1Distance} and maximizing with respect to $g(\boldx)$, we can 
obtain the tightest lower bound as
\begin{align}
\int{\left|\density(\boldx) - \density'(\boldx) \right|\mathrm{d}\boldx} \geq & \sup_{g} \int{g(\boldx)\left[\density(\boldx) - \density'(\boldx)\right]\mathrm{d}\boldx} \label{eq:MainBound} \\
& \mathrm{s.t.} \: \left|g(\boldx) \right| \leq 1, \:\: \forall \boldx. \nonumber
\end{align}
It is straightforward to verify that the above relation will be met with equality when 
\begin{align*}
g(\boldx) = \operatorname{sign}\left(\density(\boldx) - \density'(\boldx) \right). 
\end{align*}
What makes the expression in the right-hand side of Eq.\eqref{eq:MainBound} 
especially useful is that the 
probability densities occur linearly in the integral.
By replacing the integrals with sample averages and searching $g(\boldx)$ from a 
parametric family (denoted as $g_\boldalpha(\boldx)$), we can write the above as 
%
%We can estimate these terms with 
%sample averages and search $g(\boldx)$ from a parametric family (denoted as $g_\boldalpha(\boldx)$) and then write the above as
% \begin{align}
% \begin{array}{ccc}
% \boldalpha = & \argmin_{\boldalpha} & \frac{1}{n'}\sum_{j=1}^{n'}g_\boldalpha(\boldx_j') - \frac{1}{n}\sum_{i=1}g_\boldalpha(\boldx_i) \\
% & \textrm{s.t.} &   \left| g_\boldalpha(\boldx) \right| \leq 1 \:\: \forall \boldx.
% \end{array}
% \label{eq:L1ObjectiveFunction}
% \end{align}
%
\begin{align}
\begin{array}{ccl}
\boldalpha = & \displaystyle \operatorname*{arg\,min}_{\boldalpha}&  \displaystyle \frac{1}{n'}\sum_{j=1}^{n'}g_\boldalpha(\boldx_j') - \frac{1}{n}\sum_{i=1}^ng_\boldalpha(\boldx_i) \\
 & \mathrm{s.t.} &  \left| g_\boldalpha(\boldx) \right| \leq 1, \:\: \forall \boldx.
\end{array}
\label{eq:L1ObjectiveFunction}
\end{align}

\subsection{Optimization}
Here we briefly discuss how to solve the problem in Eq.~\eqref{eq:L1ObjectiveFunction}. A more detailed explanation is given in Appendix~\ref{sec:OptimizationAppendix}. 

The function in Eq.~\eqref{eq:L1ObjectiveFunction} should satisfy the constraint $|g(\boldx)| \leq 1, \: \forall \: \boldx$. We can consider a clipped version of the function that always satisfies the constraint,
\begin{align*}
\widetilde{g}(\boldx) = R(g(\boldx)),
~~\mbox{where}~~
R(z) = \begin{cases}
1 & z > 1, \\
-1 & z < -1, \\
z & \mathrm{otherwise}.
\end{cases}
\end{align*}
% This gives an objective function of 
% \begin{align}
% J(\boldalpha) =  \frac{1}{n'}\sum_{i=1}^{n'}{R\left(g_\boldalpha(\boldx_i') \right)} - \frac{1}{n}\sum_{j=1}^{n}{R\left(g_\boldalpha(\boldx_i)\right)}.
% \label{eq:MainObjectiveFunction}
% \end{align}
%
%In this section we show  that we can find a local minima of the objective function in by 
%%a straightforward application of the convex-concave procedure \cite{Yuille02theconcave-convex} (as was done in \cite{ICML:collobert:2006}). 
% Here we briefly discuss how to optimize the expression in 
% Eq.\eqref{eq:L1ObjectiveFunction}.
%A more detailed explanation is given in Appendix~\ref{sec:OptimizationAppendix}. 
%
% The function in Eq.\eqref{eq:L1ObjectiveFunction} should satisfy the constraint $\left|g(\boldx) \right| \leq 1, \:\: \forall \boldx$. We can consider a clipped version of the function that always satisfies the constraint, 
% \begin{align*}
% \widetilde{g}(\boldx) = R(g(\boldx)),
% ~~\mbox{where}~~
% R(z) = \begin{cases}
% 1 & z > 1, \\
% -1 & z < -1, \\
% z & \mathrm{otherwise}.
% \end{cases}
% \end{align*}
%A linear-in-
%The function $g_\boldalpha(\boldx)$ can be express 
We use a linear-in-parameter model, 
\begin{align*}
g(\boldx) = \sum_{\ell = 1}^b {\alpha_\ell \varphi_\ell(\boldx)},
\end{align*}
where $\varphi_\ell(\boldx)$ are the basis functions. Using the above definitions,
we can rewrite Eq.\eqref{eq:L1ObjectiveFunction} as 
\begin{align}
J(\boldalpha) =  \frac{1}{n'}\sum_{i=1}^{n'}{R\left(\sum_{\ell=1}^b{\alpha_\ell \varphi_\ell(\boldx_i')}\right)} - \frac{1}{n}\sum_{j=1}^{n}{R\left(\sum_{\ell=1}^b{\alpha_\ell \varphi_\ell(\boldx_j)}\right)} + \frac{\lambda}{2}\sum_{\ell=1}^b{\alpha_\ell^2},
\label{eq:ObjectiveFunction}
\end{align}
where $\frac{\lambda}{2}\sum_{\ell=1}^b{\alpha_\ell^2}$ is a regularization term.
Although the above is a non-convex problem, we can efficiently find a local optimal solution using the \emph{convex-concave procedure} (CCCP) \cite{Yuille02theconcave-convex} (also known as \emph{difference of convex (d.c.) programming} \cite{horst1999dc}). The CCCP procedure requires the objective function to be split into a convex and concave part, 
\begin{align*}
J(\boldalpha) = J_\mathrm{vex}(\boldalpha) + J_\mathrm{cave}(\boldalpha).
\end{align*}
The concave part is then upper-bounded as
\begin{align*}
J_\mathrm{cave}(\boldalpha) \leq \bar{J}_\mathrm{cave}(\boldalpha, \boldb, \boldc),
\end{align*}
where the bound is specified by $\boldb$ and $\boldc$ (details are given in Appendix~\ref{sec:OptimizationAppendix}). This bound is convex w.r.t. $\boldb$ and $\boldc$ if $\boldalpha$ is fixed. Using this bound, the optimization problem can then be expressed as
\begin{align*}
J(\boldalpha) \leq J_\mathrm{vex}(\boldalpha) + \bar{J}_{\textrm{cave}}(\boldalpha, \boldb, \boldc).
\end{align*}
The strategy to minimize $J(\boldalpha)$ is then to alternately minimize the right-hand side 
by minimizing w.r.t. $\boldalpha$ (keeping $\boldb$ and $\boldc$ constant) and minimize w.r.t. $\boldb$ and $\boldc$ (keeping $\boldalpha$ constant). Minimization w.r.t. $\boldalpha$ minimizes the current upper bound and minimization w.r.t. $\boldb$ and $\boldc$ corresponds to 
tightening the bound at the current point.

Minimization w.r.t. $\boldb$ and $\boldc$ can be performed by
\begin{align}
b_i = \begin{cases}
0 & \sum_{\ell = 1}^b \alpha_\ell \varphi_\ell(\boldx_i') < 1, \\
1 & \mathrm{otherwise},
\end{cases}
~~\mbox{and}~~
c_j = \begin{cases}
0 & \sum_{\ell=1}^b{\alpha_\ell \varphi(\boldx_j)} < -1, \\
1 & \mathrm{otherwise.}
\end{cases} 
\label{eq:CalculateBandC}
\end{align}
Minimization of the upper bound (assuming $\boldb$ and $\boldc$ is constant) can be performed 
by solving the following convex quadratic problem:
\begin{align}
\begin{array}{cc}
%\bar{J}(\boldalpha) & = \frac{1}{n'}\sum_{i=1}^{n'}{\xi_i' } + \frac{1}{n}\sum_{j=1}^n{\xi_j} - \sum_{\ell=1}^b \alpha_\ell\frac{1}{n'} \sum_{i=1}^{n'}{b_i\varphi_\ell(\boldx_i')} - \sum_{\ell=1}^b \alpha_\ell \frac{1}{n}\sum_{j=1}^n c_j \varphi_\ell(\boldx_j) + F(\boldalpha).  \\
% row 2 
\displaystyle \bar{J}(\boldalpha) & = \displaystyle \frac{1}{n'}\sum_{i=1}^{n'}{\xi_i' } \!+\! \frac{1}{n}\sum_{j=1}^n{\xi_j} \!-\! \sum_{\ell=1}^b\alpha_\ell\!\left(\!\frac{1}{n'} \sum_{i=1}^{n'}{b_i\varphi_\ell(\boldx_i')}  \!+\! \frac{1}{n}\sum_{j=1}^n c_j \varphi_\ell(\boldx_j)\! \right)\!   \!+\! \frac{\lambda}{2}\sum_{\ell=1}^b{\alpha_\ell^2}
\!\!\!\!\!\!\!\!\!\!\!\!\!\!\!\!
 \\
% row 3
& \displaystyle \mathrm{s.t.}\:  \xi_i' \geq 0, \: \xi_i' \geq \sum_{\ell=1}^b{\alpha_\ell \varphi_\ell(\boldx_i')} +1, \:\: \forall  i=1, \ldots, n' \\
& \displaystyle \phantom{\mathrm{s.t.}}\: \xi_j \geq 0, \: \xi_j \geq \sum_{\ell=1}^b{\alpha_\ell \varphi_\ell(\boldx_j)} - 1 \:\: \forall  j=1, \ldots, n.
\end{array}
\label{eq:UpperBoundProblem}
\end{align}
The above constrained problem can be solved with an off-the-shelf QP solver. 

Our final optimization algorithm is summarized below:
\begin{enumerate} 
  \item \emph{Initialize the starting value:} %\newline \noindent
  \begin{align*}
  \boldalpha^1 \leftarrow \displaystyle \operatorname*{arg\, min}_{\boldalpha} J_\mathrm{vex}(\boldalpha).
  \end{align*} 
  \item \emph{For $t=1, \ldots T $: } 
  \begin{enumerate}
    \item \emph{Tighten the upper-bound:} %\newline \noindent
    Obtain $\boldb$ and $\boldc$ as
    \begin{align*}
    \boldb, \boldc \leftarrow  \operatorname*{arg\, min}_{\boldb, \boldc} \bar{J}(\boldalpha^{t}, \boldb, \boldc),
    \end{align*}
    by using Eq.\eqref{eq:CalculateBandC}. 
	  \item \emph{Minimize the upper bound: } \newline \noindent
	  Set
	  \begin{align*}
	  \boldalpha^{t+1} \leftarrow \operatorname*{arg\, min}_{\boldalpha} J_\mathrm{vex}(\boldalpha) + \bar{J}_\mathrm{cave}(\boldalpha, \boldb, \boldc,)
	  \end{align*}
	  by solving the convex problem in Eq.\eqref{eq:UpperBoundProblem}.
    \end{enumerate} 
\end{enumerate}

In practice, Gaussian kernels centered at the sample points in $\calX_p$ and $\calX_{p'}$ are chosen as the basis functions. All hyper-parameters are set by cross-validation.

\section{Experiments}
\label{sec:Experiments}
We first illustrate the operation of our method and characterize the failures of other methods on various toy examples. Then we use real-world benchmark data to show the superiority of our algorithm.
%\subsection{Numerical Illustration}
%\input{toyprob1}

\subsection{Numerical Illustration}

\subsubsection{Toy Problem 1:}
We illustrate the problem and our method with a simple example. Suppose that the class-conditional densities for the two classes are given as 
\begin{align*}
p(\boldx|y=1) = \calN_\boldx\left(-\bold1_{2}, \boldI_{2\times 2}\right) 
~~\mbox{and}~~
p(\boldx|y=-1) = \calN_\boldx\left(\bold1_{2}, \boldI_{2\times 2}\right),
\end{align*}
where $\calN_\boldx(\boldmu,\boldSigma)$ denotes the normal density
with mean $\boldmu$ and covariance $\boldSigma$ w.r.t. $\boldx$.
$\bold1_{2}$ is a $2\times 1$ vector of ones and $\boldI$ is a $2\times 2$ identity matrix. We generate $2$ sets of $30$ samples
with class-priors $p(y=1)=0.3$ and $p'(y=1)=0.7$, respectively.
%  from each class according to the following class priors: 
% \begin{align*}
% \theta = 0.3, \qquad \phi = 0.6.
% \end{align*} 
The result is illustrated in Figure~\ref{fig:toyillustration}. As can be seen from this example, we are able to obtain a labeling of the classes that roughly corresponds to the true (unknown) labels of the data.
%\usepackage{graphics} is needed for \includegraphics

% \begin{figure}[t]
%         \centering
%         \begin{subfigure}[b]{0.325\textwidth}
%                 \centering
%                 \includegraphics[width=\textwidth]{toynew} 
%                \caption{}
%                 \label{fig:originalData}
%         \end{subfigure}%
%         \begin{subfigure}[b]{0.325\textwidth}
%                 \centering
%                 \includegraphics[width=\textwidth]{toyexample3}
%                 \caption{}
%                 \label{fig:labellingXp} 
%         \end{subfigure}
%         \begin{subfigure}[b]{0.325\textwidth}
%                 \centering
%                 \includegraphics[width=\textwidth]{toyexample4}
%                 \caption{}
%                 \label{fig:labellingXq} 
%         \end{subfigure}
%         \caption{Illustrative example of the class-prior change labeling method. Figure~\ref{fig:originalData} shows the samples of datasets $\calX_p$ and $\calX_{p'}$, drawn from $p(\boldx)$ and $p'(\boldx)$. Figures~\ref{fig:labellingXp} and \ref{fig:labellingXq} illustrate the discriminant applied to the datasets $\calX_{p}$ and $\calX_{p'}$. The estimated discriminant is given in blue and the optimal discriminant is in black. The true underlying class labels (which are unknown) are illustrated in red and black.  }\label{fig:toyillustration}
% \end{figure}

\subsubsection{Toy Problem 2:}
\begin{figure}[p]
        \centering
        \begin{subfigure}[b]{0.48\textwidth}
                \centering
                \includegraphics[width=\textwidth]{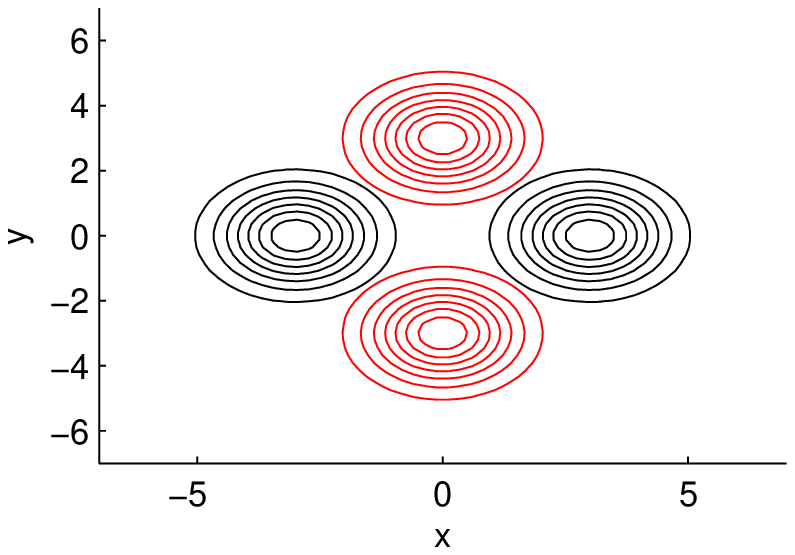} 
               \caption{Original distributions}
                \label{fig:ToyProb3Dist}
        \end{subfigure} \\
        \begin{subfigure}[b]{0.48\textwidth}
                \centering
                \includegraphics[width=\textwidth]{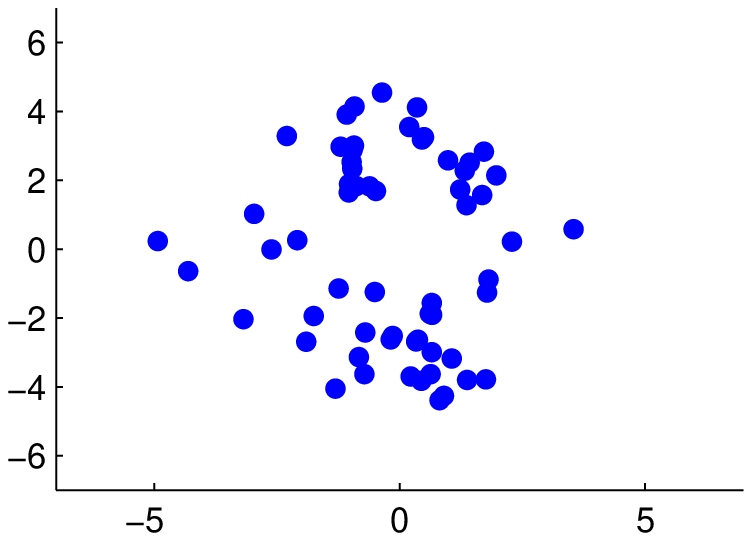}
                \caption{$\calX_{p}$}
                \label{fig:Toy3Xp} 
        \end{subfigure} 
        \begin{subfigure}[b]{0.48\textwidth}
                \centering
                \includegraphics[width=\textwidth]{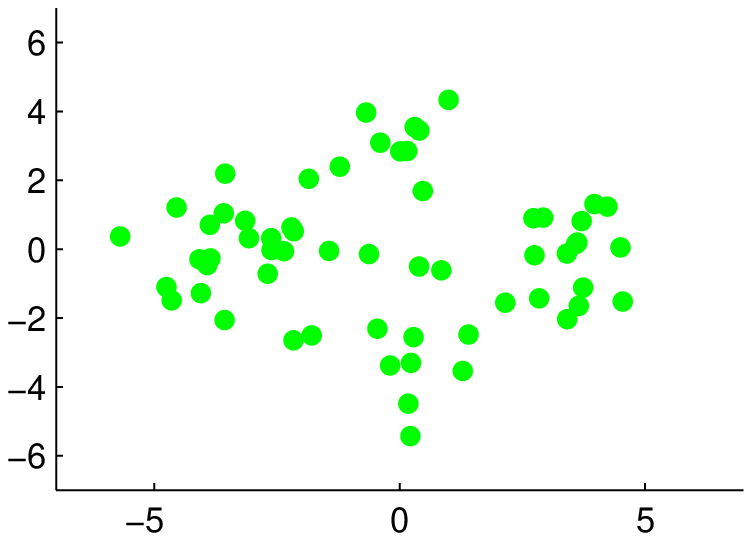}
                \caption{$\calX_{p'}$}
                \label{fig:Toy3Xq} 
        \end{subfigure} \\
        \begin{subfigure}[b]{0.48\textwidth}
                \centering
                \includegraphics[width=\textwidth]{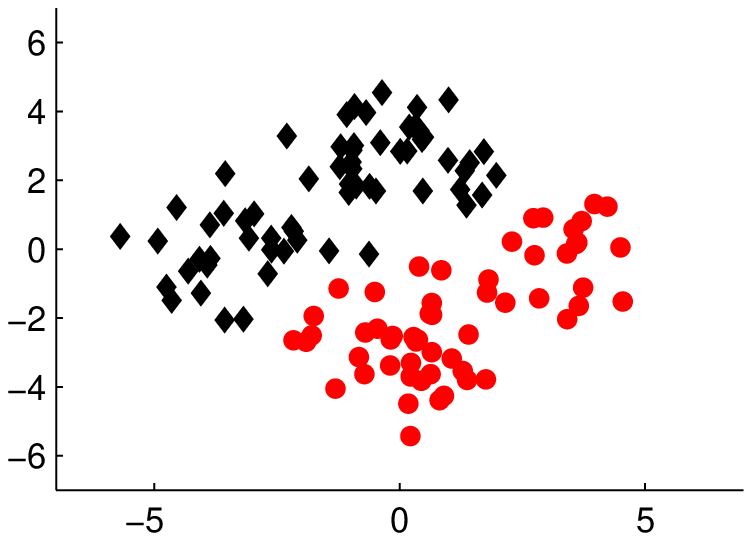} 
               \caption{K-means}
                \label{fig:toy3kmeans}
        \end{subfigure} 
        \begin{subfigure}[b]{0.48\textwidth}
                \centering
                \includegraphics[width=\textwidth]{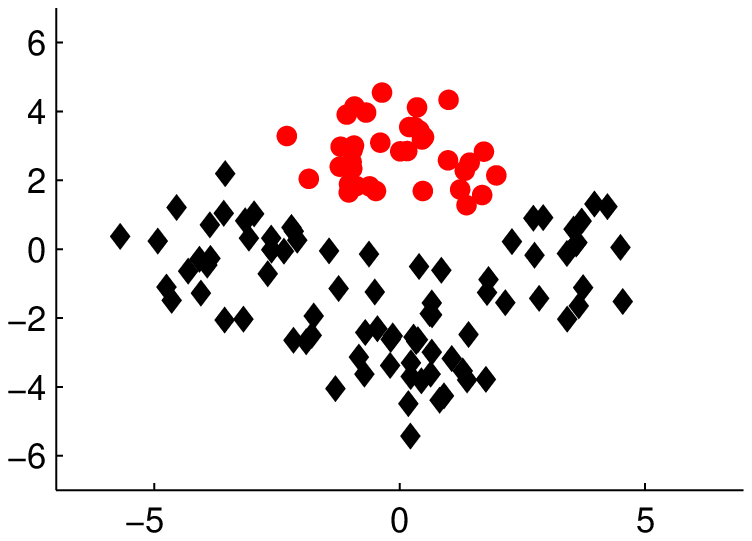}
                \caption{Spectral clustering}
                \label{fig:toy3speccluster} 
        \end{subfigure}  \\
        \begin{subfigure}[b]{0.48\textwidth}
                \centering
                \includegraphics[width=\textwidth]{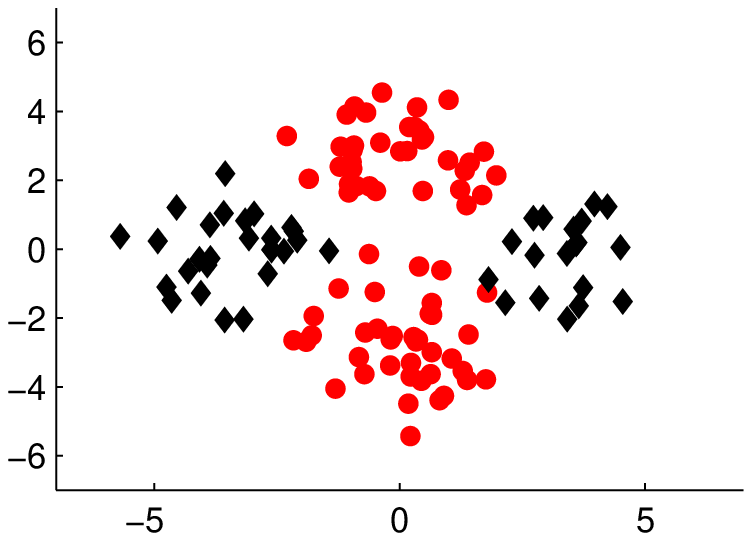}
                \caption{Direct estimation}
                \label{fig:toy3l1dd} 
        \end{subfigure}
        \caption{Illustration of within-class multimodality and clustering.}
        \label{fig:toyillustration2}
\end{figure}

%\subsubsection{Toy problem 2}
One way to obtain a labeling is to use clustering. 
The tacit assumption in clustering
is that samples in the same cluster belong the same class. 
This assumption however is not always be true,
for example, when the class conditional densities are multimodal.
Here we consider a problem with the following class conditional densities:
\begin{align*}
p(\boldx|y=1) &= \frac{1}{2}\calN_\boldx(\left[3 \: 0 \right]^\top, \boldI_{2\times 2}) + \frac{1}{2}\calN_\boldx(\left[-3 \: 0 \right]^\top, \boldI_{2\times 2}) \\
p(\boldx|y=-1) &= \frac{1}{2}\calN_\boldx(\left[0 \: 3 \right]^\top, \boldI_{2\times 2}) + \frac{1}{2}\calN_\boldx(\left[0 \: -3 \right]^\top, \boldI_{2\times 2}).
\end{align*}
The two distributions are plotted in Figure~\ref{fig:ToyProb3Dist}. We can try to obtain a class label by performing clustering on $\calX_\density \cup \calX_{\density'}$ \footnote{
If clustering is performed separately on $\calX_p$ and $\calX_{p'}$, we do not know which clusters in each dataset correspond to the clusters in the other dataset. We can also not perform clustering on one dataset and apply it to the other dataset, since most clustering methods do not give out of sample labeling. For these reasons, it makes most sense to perform clustering on the combined dataset.}. 
The results for k-means and spectral clustering, given in Figures~\ref{fig:toy3kmeans} and \ref{fig:toy3speccluster}, show that these methods fail to reveal the true labeling.
On the other hand, the proposed method still gives a reasonable result (Figure~\ref{fig:toy3l1dd}).

\subsection{Benchmark Datasets}
% \subsubsection{Setup:}
We compare our method against several competing methods on benchmark datasets.
For each experiment, we constructed the datasets $\calX_p$ and $\calX_{p'}$ by drawing $n$ and $n'$ 
samples from the positive and negative classes of the datasets according to a prior of
$p(y=1)$ and $p'(y=1)$. The labeling was then performed using these two datasets.
Since we can obtain a labeling, but cannot determine the original class labels, we cannot measure the performance using the misclassification rate directly. Assume that the label assigned for the sample $\boldx_i$ is 
\begin{align*}
l_i = \begin{cases}
-1 & p(\boldx_i) - q(\boldx_i) < 0 \\
1 & \textrm{otherwise}. 
\end{cases} 
\end{align*}
The misclassification rate (MCR) assuming that the current labels are correct is 
\begin{align*}
\textrm{MCR} := \frac{1}{n}\sum_{i:l_j \neq y_i} {1} + \frac{1}{n'}\sum_{j:l_j' \neq y_i'} {1}.
\end{align*}
The misclassification rate assuming that the labels are the opposite is $1-\textrm{MCR}$. 
We define the \emph{labeling error rate} (LER) as
\begin{align*}
\mathrm{LER} := \min\left(\textrm{MCR}, 1-\textrm{MCR}\right).
\end{align*} 
Note that this definition is somewhat more optimistic than using the misclassification rate.
The smaller the dataset is, the lower the error would be for randomly 
assigning labels to samples: 
The expected LER for randomly assigning labels to samples (with equal probability) is
\begin{align*}
\frac{1}{2^{n+n'}(n+n')}\sum_{i=0}^{n+n'}{\min{\left(i, n+n'-i \right)}\binom{n+n'}{i}}.
\end{align*}  
For $n+n' = 40, 60, 80$, the expected labeling error rate is $0.437,  0.449, 0.456$.

We compared the following methods:
\begin{itemize}
    \item \textbf{Direct Sign Density Difference (DSDD) Estimation} (proposed): Directly estimate $\operatorname{sign}{ (\density(\boldx) - \density'(\boldx))} $ using the method described in Section~\ref{sec:DirectEstimationSign}. Hyperparameters are selected via cross validation. 
  \item \textbf{Least-Squares Density Difference (LSDD) Estimation}: Estimate $\operatorname{sign}\left[\density(\boldx) - \density'(\boldx)\right]$ by estimating $\density(\boldx) - \density'(\boldx)$ using the least squares fitting method \cite{DensityDifferenceNECO}.
Hyperparameters are selected via cross validation. 
  \item \textbf{Kernel Density Estimation (KDE):} Estimate $\operatorname{sign}\left[\density(\boldx) - \density'(\boldx)\right]$ by estimating the densities $\density(\boldx)$ and $\density'(\boldx)$ with kernel density estimation (KDE). Hyperparameters are selected using least-squares cross validation.
  \item \textbf{K-Means (KM)}: Cluster the data into two clusters using the K-means algorithm. 
  \item \textbf{Spectral Clustering (SC)}: Cluster the data into two clusters using the spectral clustering algorithm \cite{Shi-Malik-SpectralClustering}. The affinity matrix was constructed with $7$ nearest neighbors.
  \item \textbf{Squared-loss Mutual Information based Clustering (SMIC) } : Cluster the data
  according to the SMIC method \cite{ICML:Sugiyama+etal:2011}. SMIC was chosen since it provides model selection, avoiding the need for subjective parameter tuning.
\end{itemize}

% \subsubsection{Results and Discussion:}
We compare the performance of the methods by varying the class balance.
Two class balances were selected: one with a large difference between the classes ($p(y=1) = 0.2$ and $p'(y=1)=0.8$) and one with a small difference between the two priors ($p(y=1) = 0.35$ and $p'(y=1)=0.65$). 
The average labeling error rate and standard deviation of the two experiments, with $\left| \calX_{p}\right| = \left|\calX_{p'} \right| = 40$ is given in Tables~\ref{tab:Results-02-08} and \ref{tab:Results-035-065}.

\begin{table}[t]
\centering
\small
\caption{Labeling error rate for experiments with a class prior of $p(y=1)=0.2$ and $p'(y=1)=0.8$. The size of each dataset was $\left| \calX_{p}\right| = 40$ and $\left| \calX_{p'}\right| = 40$. The best method in terms of the mean error and
comparable methods according to the two-sided paired t-test at the significance level 5\%
are specified by bold face. The standard deviation of the labeling error rate is given in brackets.}

% 
% This table was automatically generated by code
% c@{1}c
% Np 40 Nq 40
% Theta 0.2 Phi 0.8
\begin{tabular}{p{1.60cm}|c@{\hskip 1pt}c@{\hskip 8pt}c@{\hskip 1pt}c@{\hskip 8pt}c@{\hskip 1pt}c@{\hskip 8pt}c@{\hskip 1pt}c@{\hskip 8pt}c@{\hskip 1pt}c@{\hskip 8pt}c@{ \hskip 1pt}c}  
\toprule  
 { \textbf{Dataset}}& \multicolumn{2}{c}{\textbf{DSDD} } & \multicolumn{2}{c}{\textbf{LSDD}}  & \multicolumn{2}{c}{\textbf{KDE}}  & \multicolumn{2}{c}{\textbf{KM}}  & \multicolumn{2}{c}{\textbf{SC}}  & \multicolumn{2}{c}{\textbf{SMIC} }  \\
 \midrule 
% Row 1 Dataset australian
australian & \textbf{.142} & (.045)  & {.174} & (.110)  & {.211} & (.126)  & {.266} & (.147)  & {.381} & (.033)  & {.303} & (.103) \\ 
 	 % 1558
% Row 2 Dataset banana
banana & {.179} & (.097)  & \textbf{.170} & (.070)  & {.237} & (.147)  & {.431} & (.068)  & {.427} & (.141)  & {.424} & (.141) \\ 
 	 % 1558
% Row 3 Dataset diabetes
diabetes & {.246} & (.122)  & \textbf{.223} & (.079)  & \textbf{.226} & (.051)  & {.372} & (.080)  & {.380} & (.094)  & {.370} & (.131) \\ 
 	 % 1577
% Row 4 Dataset german
german & {.268} & (.059)  & {.281} & (.127)  & \textbf{.211} & (.051)  & {.437} & (.114)  & {.448} & (.128)  & {.439} & (.052) \\ 
 	 % 1558
% Row 5 Dataset heart
heart & \textbf{.176} & (.051)  & \textbf{.174} & (.047)  & {.211} & (.074)  & {.261} & (.131)  & {.310} & (.032)  & {.327} & (.107) \\ 
 	 % 1558
% Row 6 Dataset image
image & \textbf{.198} & (.078)  & {.206} & (.047)  & \textbf{.201} & (.049)  & {.385} & (.093)  & {.351} & (.119)  & {.384} & (.135) \\ 
 	 % 1539
% Row 7 Dataset ionosphere
ionosphere & \textbf{.157} & (.059)  & {.184} & (.106)  & {.194} & (.123)  & {.329} & (.145)  & {.319} & (.113)  & {.311} & (.174) \\ 
 	 % 1558
% Row 8 Dataset saheart
saheart & {.310} & (.093)  & \textbf{.205} & (.048)  & {.238} & (.113)  & {.422} & (.121)  & {.395} & (.113)  & {.384} & (.072) \\ 
 	 % 1558
% Row 9 Dataset thyroid
thyroid & \textbf{.102} & (.052)  & {.121} & (.116)  & {.207} & (.074)  & {.328} & (.113)  & {.326} & (.109)  & {.305} & (.074) \\ 
 	 % 1558
% Row 10 Dataset twonorm
twonorm & {.044} & (.085)  & {.051} & (.072)  & {.200} & (.028)  & \textbf{.036} & (.054)  & {.043} & (.069)  & {.048} & (.071) \\ 
 	 % 1558
 
 \bottomrule \end{tabular}

%\caption{$N=30 \: 0.3 \: 0.6$}
\label{tab:Results-02-08}
\end{table}  
\begin{table}[t]
\centering
\small
\caption{Labeling error rate for experiments with a class prior of $p(y=1)=0.35$ and $p'(y=1)=0.65$. The size of each dataset was $\left| \calX_{p}\right| = 40$ and $\left| \calX_{p'}\right| = 40$. The best method in terms of the mean error and
comparable methods according to the two-sided paired t-test at the significance level 5\%
are specified by bold face. The standard deviation of the labeling error rate is given in brackets.}

%\begin{tabular}{p{1.55cm}p{1.7cm}p{1.7cm}p{1.7cm}p{1.7cm}p{1.7cm}l}
\begin{tabular}{p{1.60cm}|c@{\hskip 1pt}c@{\hskip 8pt}c@{\hskip 1pt}c@{\hskip 8pt}c@{\hskip 1pt}c@{\hskip 8pt}c@{\hskip 1pt}c@{\hskip 8pt}c@{\hskip 1pt}c@{\hskip 8pt}c@{ \hskip 1pt}c}  
\toprule  
 { \textbf{Dataset}}& \multicolumn{2}{c}{\textbf{DSDD} } & \multicolumn{2}{c}{\textbf{LSDD}}  & \multicolumn{2}{c}{\textbf{KDE}}  & \multicolumn{2}{c}{\textbf{KM}}  & \multicolumn{2}{c}{\textbf{SC}}  & \multicolumn{2}{c}{\textbf{SMIC} }  \\
 \midrule 
% Row 1 Dataset australian
australian & \textbf{.244} & (.116)  & {.259} & (.088)  & {.355} & (.104)  & {.265} & (.080)  & {.376} & (.065)  & {.308} & (.107) \\  
 	 % 1577
% Row 2 Dataset banana
banana & \textbf{.338} & (.094)  & \textbf{.339} & (.100)  & {.365} & (.067)  & {.433} & (.049)  & {.427} & (.069)  & {.424} & (.070) \\ 
 	 % 1558
% Row 3 Dataset diabetes
diabetes & \textbf{.340} & (.075)  & {.361} & (.124)  & {.345} & (.034)  & {.373} & (.063)  & {.380} & (.048)  & {.371} & (.114) \\ 
 	 % 1558
% Row 4 Dataset german
german & {.375} & (.042)  & {.380} & (.093)  & \textbf{.354} & (.057)  & {.437} & (.024)  & {.445} & (.057)  & {.438} & (.041) \\ 
 	 % 1558
% Row 5 Dataset heart
heart & {.270} & (.133)  & \textbf{.247} & (.084)  & {.354} & (.052)  & {.264} & (.059)  & {.315} & (.081)  & {.327} & (.089) \\ 
 	 % 1558
% Row 6 Dataset image
image & \textbf{.331} & (.078)  & {.350} & (.067)  & {.350} & (.039)  & {.384} & (.031)  & {.354} & (.049)  & {.382} & (.050) \\ 
 	 % 1539
% Row 7 Dataset ionosphere
ionosphere & \textbf{.291} & (.099)  & {.356} & (.066)  & {.345} & (.048)  & {.330} & (.070)  & {.322} & (.058)  & {.314} & (.107) \\ 
 	 % 1539
% Row 8 Dataset saheart
saheart & {.378} & (.093)  & \textbf{.353} & (.057)  & {.363} & (.066)  & {.419} & (.082)  & {.395} & (.022)  & {.385} & (.040) \\ 
 	 % 1558
% Row 9 Dataset thyroid
thyroid & \textbf{.227} & (.098)  & {.251} & (.087)  & {.302} & (.022)  & {.326} & (.061)  & {.329} & (.047)  & {.307} & (.076) \\ 
 	 % 1520
% Row 10 Dataset twonorm
twonorm & {.164} & (.188)  & {.153} & (.121)  & {.352} & (.096)  & \textbf{.036} & (.053)  & {.042} & (.122)  & {.049} & (.120) \\ 
 	 % 1558
 
 \bottomrule \end{tabular}

%\caption{$N=30 \: 0.3 \: 0.6$}
\label{tab:Results-035-065}
\end{table}

From the results we see that methods which follow the approach proposed in Section~\ref{sec:ProblemFormulation}
of estimating the sign of the density difference (i.e., DSDD, LSDD, and KDE)
generally work better than methods using the cluster structure of the 
data (i.e., KM, SC and SMIC).
% Among them, the proposed DSDD compares favorably with other approaches.
The thyroid dataset lends itself to interpretation of why these methods work better. 
The labels in the thyroid dataset correspond to healthy and diseased. The diseased label is caused by either a hyper-functioning or hypo-functioning thyroid. These two underlying
causes cause within-class multimodality which may cause clustering-based methods to fail.

Among the methods which estimate the sign of the 
density difference, we see that DSDD generally performs better than 
LSDD and LSDD in turn performs better than KDE. 
This is as expected since KDE solves a more general problem than LSDD, and LSDD solves a more general problem than DSDD.
This pattern is even more pronounced on the more difficult case where the 
class balances are close to each other (Table~\ref{tab:Results-035-065}).

\section{Conclusion}
The problem of unsupervised labeling of two unbalanced datasets was considered. We first showed that this problem can be solved if two unlabeled datasets having different class balances are available. The solution can be obtained by estimating of the sign of the difference between probability densities. We introduced a method to directly estimate the sign of the density difference and avoid density estimation. The method was shown on various datasets to outperform competing methods that either estimate the density difference or use the cluster structure of the data. 

Because the sign of density difference corresponds to the Bayes optimal 
classifier under equal class balance,
it may be estimated by any classifier that separates $\calX_p$ and $\calX_{p'}$.
Following this idea, we tested the \emph{support vector machine} (SVM)
for estimating the sign of density difference.
However, this did not work well due to the high overlap of $\calX_p$
and $\calX_{p'}$---both
the datasets are mixtures of two classes, only with different mixing ratios.

From this classification point of view,
we can actually see that our objective function \eqref{eq:ObjectiveFunction}
corresponds to the \emph{robust SVM} \cite{Shawe-Taylor:2004:KMP:975545}
that minimizes the ramp loss (a clipped hinge loss).
Thanks to the robustness brought by the ramp loss,
the overlapped datasets $\calX_p$ and $\calX_{p'}$ can be separated
more reliably, and thus we obtained good estimation of the sign of
density difference.

Furthermore, this view conversely shows that the robust SVM is
actually a suitable
classification method because it directly estimates the Bayes optimal
classifier,
the sign of density difference.
Labeling and classification are different problems,
but one can actually give insight into the other.
In the future work,
we will further investigate the relation between
labeling and classification.

%In practice, the sign of density difference can be estimated from any classifier.  In our preliminary experiments however, the support vector machine (SVM) did not work well due to the overlap of datasets caused by a change in class priors. Our objective function corresponds to robust SVM \cite{Shawe-Taylor:2004:KMP:975545}, which is natural for this type of overlapped dataset. Robust SVM, in contrast to SVM, actually estimates the \emph{sign} directly.
%Investigating this relationship in the future would be promising. 

% \vspace{2cm}
% The sign of density difference can be estimated by any classifier. 
% In our preliminary experiments, SVM did not work due to 
% the complete overlap of datasets, (which is the nature of this problem, since prior change occur).
% Our objective function corresponds to robust SVM, which 
% is natural for this type of overlapped dataset. 
% Robust SVM is actually estimating the sign directly. 
% %
% \hrule
% The problem of unsupervised labeling of two unlabeled datasets was considered. 
% We first showed that this problem can be solved if two unlabeled datasets having
% different class balances are available. The solution can be obtained by estimation 
% of the sign of the difference between probability densities. We introduced
% a method to directly estimate the sign between the density difference by avoiding 
% density estimation.
% The efficacy of the proposed method compared to the naive method of clustering was shown
% on various toy examples and various real world datasets. 

\bibliographystyle{splncs}
\bibliography{bib}

\appendix

\section{Optimization}
\label{sec:OptimizationAppendix}
This section outlines the optimization of Eq.~\eqref{eq:ObjectiveFunction} using the convex concave procedure\cite{Yuille02theconcave-convex}. The non-convex function $R(z)$ can be re-written as
\begin{align*}
R(z) = C_{-1}(z) - C_{1}(z)-1, \:\:\textrm{where }\:\:C_{\epsilon}(z) = \max(0, z-\epsilon).
\end{align*} 
The convex part of the objective function can then be expressed as
\begin{align*}
J_{\mathrm{vex}}(\boldalpha) & = \frac{1}{n'}\sum_{i=1}^{n'}C_{-1}\left(\sum_{\ell=1}^{b}\alpha_\ell \varphi_\ell(\boldx_i') \right) + \frac{1}{n}\sum_{j=1}^nC_{1}\left(\sum_{\ell=1}^b \alpha_\ell \varphi_\ell(\boldx_j)  \right) + \frac{\lambda}{2}\sum_{\ell=1}^b{\alpha_\ell^2},
\end{align*}
and the concave part as
\begin{align*}
J_{\mathrm{cave}}(\boldalpha) & = -\frac{1}{n'}\sum_{i=1}^{n'}C_{1}\left(\sum_{\ell=1}^b \alpha_\ell \varphi_\ell (\boldx_i') \right) - \frac{1}{n}\sum_{j=1}^n {C_{-1}\left(\sum_{\ell=1}^b \alpha_\ell \varphi_\ell (\boldx_j) \right)}.
\end{align*}
The following self-evident relation can be used to bound the concave part
\begin{align*}
tz - \varphi(t) & \leq \sup_{y\in \mathbbR} yz - \varphi(y) \\
\Rightarrow \varphi(t) & \geq tz - \varphi^\ast(z),
%\label{eq:LFBound}
\end{align*}
where 
\begin{align*}
\varphi^\ast(z) = \sup_{y\in \mathbbR} yz - \varphi(y)
\end{align*}
is known as the \emph{convex conjugate}. The convex conjugate of the 
function $C_\epsilon(z)$ is
\begin{align*}
C_\epsilon^\ast(z) = \begin{cases}
\infty & z < 0 \\
\epsilon z & 0 \leq z \leq 1 \\
\infty & z > 0.
\end{cases}
\end{align*}
This gives an upper bound on the concave function as
\begin{align*}
\bar{J}_{\mathrm{cave}}(\boldalpha, \boldb, \boldc) & =  
% first term
\frac{1}{n'}\sum_{i=1}^{n'}{\left(C_{1}^\ast(b_i) - b_i\sum_{\ell=1}^b \alpha_\ell \varphi_\ell(\boldx_i')\right)} 
% second term 
+ \frac{1}{n}\sum_{j=1}^n{\left(C_{-1}^\ast(c_j) - c_j\sum_{\ell=1}^b \alpha_\ell \varphi_\ell(\boldx_j) \right)},
\end{align*}
where $\boldb = \left[b_1 \: b_2 \: \ldots \: b_{n'} \right]$ and $\boldc = \left[c_1 \: c_2 \: \ldots \: c_{n} \right]$ specify the bound.

\subsection{Tightening the bound}
The bound can be tightened around $\boldalpha$ by minimizing $J_{\textrm{cave}}(\boldalpha, \boldb, \boldc)$ w.r.t. $\boldb$ and $\boldc$. To ensure that we have a non-trivial bound, 
we can explicitly write the conjugate as constraints,
\begin{align*}
\begin{array}{cl}
\displaystyle \bar{J}_{\mathrm{cave}}(\boldalpha, \boldb, \boldc) & = \displaystyle \frac{1}{n'}\sum_{i=1}^{n'}{b_i \left(1- \sum_{\ell=1}^b \alpha_\ell \varphi_\ell(\boldx_i') \right) }
+ \frac{1}{n}\sum_{j=1}^n c_j\left(-1 - \sum_{\ell=1}^b \alpha_\ell \varphi_\ell(\boldx_j) \right) \\ 
% line 2 
& \phantom{=} \mathrm{s.t. } \:\: 0 \leq b_i \leq 1, 0 \leq c_j \leq 1.
\end{array}
\end{align*}
The above optimization problem is separable in all unknowns, and the optimal value 
can be obtained by Eq.~\eqref{eq:CalculateBandC}. 

\subsection{Minimizing the upper bound}
The upper bound of the objective function with $\boldb$ and $\boldc$ is 
\begin{align*}
J_{\mathrm{vex}}(\boldalpha) + \bar{J}_\textrm{cave}(\boldalpha, \boldb, \boldc).  
\end{align*}
By replacing each function $C_\epsilon(z)$ with a slack variable $\xi_i$, and the constraint
\begin{align*}
\xi_i \geq 0, \: \xi_i \geq z - \epsilon, 
\end{align*}
we obtain the objective function in Eq.~\eqref{eq:UpperBoundProblem}

\section{Least-squares estimation of the density difference}
\label{sec:AppendixLSDD}
In \cite{DensityDifferenceNECO} it was proposed to directly estimate the 
density difference by fitting a model $g(\boldx)$ to the true density difference $f(\boldx)$ under a square loss:
\begin{align*}
  \argmin_{\diffmodel}
  \frac{1}{2}\int\Big(\diffmodel(\boldx)-\left[\density(\boldx) - \density'(\boldx) \right]\Big)^2
  \mathrm{d}\boldx.
%  \label{boldtheta-ast}
\end{align*}
The density difference was modeled by a linear-in-parameter model $\diffmodel(\boldx)$:
\begin{align}
  \diffmodel(\boldx)=\sum_{\ell=1}^{\nparam}\theta_\ell\psi_\ell(\boldx)
  =\boldtheta^\top\boldpsi(\boldx), 
  \label{linear-model}
\end{align} 
where $\nparam$ denotes the number of basis functions,
$\boldpsi(\boldx)=(\psi_1(\boldx),\ldots,\psi_{\nparam}(\boldx))^{\top}$ is
a $\nparam$-dimensional basis function vector,
$\boldtheta=(\theta_1,\ldots,\theta_{\nparam})^\top$
is a $\nparam$-dimensional parameter vector,
and $^\top$ denotes the transpose.
%In practice, we use the following non-parametric Gaussian kernel model as $\diffmodel(\boldx)$:
A Gaussian kernel model is used to model the density difference:
\begin{align*}
  \diffmodel(\boldx)
  =\sum_{\ell=1}^{\nsample+\nsample'}\theta_\ell
  \exp\left(-\frac{\|\boldx-\boldc_\ell\|^2}{2\sigma^2}\right),
%  \label{Gaussian-kernel-model}
\end{align*}
where  
$
  (\boldc_1,\ldots,\boldc_{\nsample},\boldc_{\nsample+1},\ldots,\boldc_{\nsample+\nsample'})
  :=(\boldx_1,\ldots,\boldx_{\nsample},\boldx'_1,\ldots,\boldx'_{\nsample'})
$
are Gaussian kernel centers.
For the model in Eq.~\eqref{linear-model},
the optimal parameter $\boldtheta^\ast$ is given by
\begin{align*}
  \boldtheta^\ast&:=\argmin_{\boldtheta}
  \frac{1}{2}\int\Big(\diffmodel(\boldx)-\left[ \density(\boldx) - \density'(\boldx)\right] \Big)^2
  \mathrm{d}\boldx\\
  &\phantom{:}=\argmin_{\boldtheta}
  \left[
    \frac{1}{2}\int\diffmodel(\boldx)^2\mathrm{d}\boldx
    -\int\diffmodel(\boldx)\left[ \density(\boldx) - \density'(\boldx)\right]\mathrm{d}\boldx
  \right]\\
  &\phantom{:}=\argmin_{\boldtheta}
  \left[
    \frac{1}{2}\boldtheta^\top\boldH\boldtheta-\boldh^\top\boldtheta
  \right]\\
  &\phantom{:}=\boldH^{-1}\boldh,
\end{align*}
where $\boldH$ is the $\nparam\times\nparam$ matrix
and $\boldh$ is the $\nparam$-dimensional vector defined as
\begin{align*}
 \boldH&:=\int\boldpsi(\boldx)\boldpsi(\boldx)^{\top}\mathrm{d}\boldx,\\
 \boldh&:=\int\boldpsi(\boldx)\density(\boldx)\mathrm{d}\boldx
-\int\boldpsi(\boldx)\density'(\boldx)\mathrm{d}\boldx.
% \boldE_{\density}[\boldpsi]-\boldE_{\density'}[\boldpsi].
\end{align*}
For the Gaussian kernel model, the integral in $\boldH$ can be computed analytically as
\begin{align*}
  H_{\ell,\ell'}&=\int\exp\left(-\frac{\|\boldx-\boldc_\ell\|^2}{2\sigma^2}\right)
  \exp\left(-\frac{\|\boldx-\boldc_{\ell'}\|^2}{2\sigma^2}\right)\mathrm{d}\boldx\\
  &=(\pi\sigma^2)^{\inputdim/2}
  \exp\left(-\frac{\|\boldc_{\ell}-\boldc_{\ell'}\|^2}{4\sigma^2}\right),
%   h_{\ell}&=\int\exp\left(-\frac{\|\boldx-\boldc_\ell\|^2}{2\sigma^2}\right)
%   \density(\boldx)\mathrm{d}\boldx
%   -\int\exp\left(-\frac{\|\boldx-\boldc_\ell\|^2}{2\sigma^2}\right)
%   \density'(\boldx)\mathrm{d}\boldx,
\end{align*}
where $\inputdim$ is the dimensionality of $\boldx$.

Replacing the expectations in $\boldh$ by empirical estimators
and adding an $\ell_2$-regularizer to the objective function,
we arrive at the following optimization problem:
\begin{align}
\boldthetah:=\argmin_{\boldtheta}
\left[\frac{1}{2}\boldtheta^\top\boldH\boldtheta-\boldhh^\top\boldtheta + \frac{1}{2}\lambda\boldtheta^\top\boldtheta\right],
\label{LSDD}
\end{align}
where $\lambda$ ($\ge0$) is the regularization parameter
and $\boldhh$ is the $\nparam$-dimensional vector defined as
\begin{align*}
  \boldhh&=\frac{1}{\nsample}\sum_{i=1}^{\nsample}\boldpsi(\boldx_i)
  -\frac{1}{\nsample'}\sum_{j=1}^{\nsample'}\boldpsi(\boldx'_{j}).
\end{align*}
Taking the derivative of the objective function in Eq.\eqref{LSDD}
and equating it to zero,
we can obtain the solution $\boldthetah$ analytically as
\begin{align*}
  \boldthetah=\left(\boldH+\lambda\boldI_{\nparam}\right)^{-1}\boldhh,
\end{align*}
where $\boldI_{\nparam}$ denotes 
the $\nparam$-dimensional identity matrix.
%
%Finally, a density-difference estimator $\diffh(\boldx)$ is given as
Finally, the density difference estimator is 
\begin{align*}
  \diffh(\boldx)
  =\boldthetah^\top\boldpsi(\boldx).
%\label{LSDD-fhat} 
\end{align*}

% comparison with the hinge loss
%\input{hingelosscomparison}

%\newpage

\end{document}